\documentclass[
]{ceurart}

\usepackage{listings}
\usepackage{float}
\usepackage{graphicx}
\usepackage{subcaption}
\usepackage{mathtools}

\usepackage[ruled, vlined, commentsnumbered, longend]{algorithm2e}

\begin{document}

\copyrightyear{2024}
\copyrightclause{Copyright for this paper by its authors.
  Use permitted under Creative Commons License Attribution 4.0
  International (CC BY 4.0).}

\conference{Dubrovnik'24: Workshop of Edge AI meets Swarm Intelligence,
  September 18, 2024, Dubrovnik, Croatia}

\title{A comparison of extended object tracking with multi-modal sensors in indoor environment}

% \tnotetext[1]{You can use this document as the template for preparing your
%   publication. We recommend using the latest version of the ceurart style.}

\author[1]{Jiangtao Shuai}[%
% orcid=0000-0002-0877-7063,
email=jiangtao.shuai@tu-berlin.de,
% url=https://yamadharma.github.io/,
]

\cormark[1]
\address[1]{Open Distributed Systems, Technical University of Berlin,
  Einsteinufer 25, 10587 Berlin, Germany}
  \author[2]{Martin Baerveldt}
\address[2]{Dept. of Engineering Cybernetics, Norwegian University of Science and Technology, Trondheim, Norway}
% \address[2]{Joint Institute for Nuclear Research,
%   6 Joliot-Curie, Dubna, Moscow region, 141980, Russian Federation}
\author[1]{Manh Nguyen-Duc}
\author[1]{Anh Le-Tuan}
\author[1]{Manfred Hauswirth}
\author[1]{Danh Le-Phuoc}

\cortext[1]{Corresponding author.}
% \fntext[1]{These authors contributed equally.}

\begin{abstract}
  This paper presents a preliminary study of an efficient object tracking approach, comparing the performance of two different 3D point cloud sensory sources: LiDAR and stereo cameras, which have significant price differences. 
  In this preliminary work, we focus on single object tracking. We first developed a fast heuristic object detector that utilizes prior information about the environment and target. The resulting target points are subsequently fed into an extended object tracking framework, where the target shape is parameterized using a star-convex hypersurface model.
  Experimental results show that our object tracking method using a stereo camera achieves performance similar to that of a LiDAR sensor, with a cost difference of more than tenfold.
\end{abstract}


\maketitle

\section{Introduction}

In a smart factory, a surveillance system that tracks mobile robots can increase the shared situational awareness of collaborative manufacturing applications. To facilitate adaptability and scalability, it is essential to develop low-cost solutions that can run efficiently on edge devices. Different from the traditional tracking scenario, where an object only generates a single measurement, 
the modern exteroceptive sensors adopted in smart manufacturing applications can provide multiple 3D measurements for each object, increasing the complexity of the tracking problem \cite{granstrom2023tutorial}. Furthermore, 
compared to the tracking approaches based on SORT \cite{bewley2016sort} that focus on the bounding boxes of targets, Extended Object Tracking (EOT) directly uses sensor measurements to recursively estimate not only the position and kinematics but also the spatial extent of the targets, which especially benefits the surveillance system for mobile robots in factory environments. 

EOT has been used in the maritime and automotive domains \cite{pandharipande2023sensing, baerveldt2023extended, cao2018extended, xia2021learning, ruud2018lidar}, where sensors like LiDAR and radar generate 3D point clouds with detection ranges from hundreds of meters to kilometers. Stereo cameras, e.g., Intel Realsense, which are much cheaper, can only generate 3D point clouds within a limited detection range (usually up to 10 meters), making them unsuitable for EOT in these domains. However, for mobile robot tracking on the factory floor, this detection range limitation is no longer an issue.
To the best of our knowledge, no related work on analyzing EOT using stereo camera point clouds has been reported to date.

In this paper, we compared the tracking performance between LiDAR and stereo camera sensors in an indoor environment. For simplification, a single mobile robot is used as the target. We first developed an efficient object detection method based on the Density-Based Spatial Clustering of Applications with Noise (DBSCAN) \cite{ester1996dbscan}, utilizing prior information about the environment and the target robot. Then, based on the detection results, EOT is implemented as described in \cite{baerveldt2023extended}. The performances are evaluated using two different sensors: a 3D LiDAR Blickfeld Cube 1 \cite{blickfeld}, which costs more than 4000 euros, and a stereo camera Intel RealSense D435i \cite{d435i}, which costs less than 400 euros.

% TODO: cite the sensor product website page

\section{Background and Related work}
In this section, we first introduce the differences between 3D point clouds generated by LiDAR and stereo cameras. Then, we give a brief description of the Gaussian Process (GP) target model \cite{wahlstrom2015extended} working in EOT; more details can be found in \cite{baerveldt2023extended}.

\subsection{3D point cloud from LiDAR and camera}
3D point clouds are widely used in perception tasks such as object detection, tracking, and segmentation \cite{guo2020deep}. 
% In contrast to 2D data representation, the 3D point clouds can include the objects' natural spatial information. 
Two common sensory sources for 3D point clouds are LiDAR and stereo cameras \cite{liu2019deep}.

A LiDAR sensor periodically emits laser beams and captures the reflections from the surfaces of the objects it hits. The depth information is calculated based on the time of flight. Additionally, the LiDAR sensor can capture reflection intensity information, which represents the properties of the object's surface material. In mainstream LiDAR sensors, such as those from Velodyne, the accuracy of generated points within the operating range can reach ±2 cm \cite{kelly2022accuracy}.

On the other hand, a stereo camera uses two lenses to generate two individual images. Based on the focal length, pixel disparity, and the baseline between the two lenses, the depth information is calculated. Due to this method of obtaining depth information, the valid depth range from a stereo camera is usually shorter than that of a LiDAR and is more sensitive to occlusions. For instance, the Stereolabs ZED can capture depth information up to 20 meters \cite{zed2024}. However, compared to LiDAR’s point cloud, the one from a stereo camera provides RGB data for each point, offering richer information than intensity and easily integrating with existing image-based feature extraction algorithms. In this paper, 3D point clouds are collected using both LiDAR and a stereo camera, and only spatial data in Cartesian coordinates are used.

\subsection{Extended object tracking}
The aim of EOT is to model measurements with the spatial distribution of the sensor data over the extent of the target and then use Bayesian filtering to predict and update the extended state \cite{pandharipande2023sensing}.
With a random hypersurface model, which models star-convex shapes by parameterizing the shape contour \cite{granstrom2016extended}, an EOT framework consists of the following elements:
\begin{itemize}
    \item the augmented state vector at time step $k$: $\boldsymbol{x_{k}} = [\boldsymbol{p_{k}^{c}}, \boldsymbol{\dot{p_{k}^{c}}}, \boldsymbol{p_{k}^{f}}]^{T}$, where the vector $\boldsymbol{p_{k}^{c}}$ represents the centric position and heading of the target, $\boldsymbol{\dot{p_{k}^{c}}}$ is the kinematic states vector in terms of velocities, $\boldsymbol{p_{k}^{f}}$ represents the parameterization vector of the extent. 
    \item a state transition model that describes the temporal evolution of the states: $f( \boldsymbol{x_{k+1}} \mid \boldsymbol{x_{k}})$.
    \item the available measurements up to time step $k$: $\mathbb{Z}_{1:k} \coloneqq \{ Z_{1}, \ldots, Z_{k} \} $, where $Z_{k} \coloneqq \{ \boldsymbol{z_{k}^{1}}, \ldots , \boldsymbol{z_{k}^{m}} \}$ represents all measurements at time step $k$.
    \item a probabilistic measurement model $f( Z_{k}  \mid \boldsymbol{{x_{k}}})$ gives the likelihood that the target object with state $\boldsymbol{x_{k}}$ randomly generates $m$ measurements, where $m$ is usually assumed to be sampled from a Poisson distribution.
    % \item a probabilistic measurement model $f( Z_{k}^{n}  \mid \boldsymbol{{x_{k}}})$ gives the likelihood that the target object with state $\boldsymbol{x_{k}}$ randomly generates $n$ measurements, where $Z_{k}^{n} \subseteq Z_{k}$.
    \item a spatial distribution model that assumes each measurement is a sample of a random variable conditioned on the target states: $f( \boldsymbol{z_{k}^{j}} \mid \boldsymbol{x_{k}})$, $\forall j \in \{ 1, \ldots, m \} $.
    \item a Bayesian estimator includes prediction $f(\boldsymbol{x_{k}} \mid \mathbb{Z}_{1:k-1})$ and update $f(\boldsymbol{x_{k}} \mid \mathbb{Z}_{1:k}) = \frac{ f( Z_{k} \mid \boldsymbol{x_{k}}) \cdot f(\boldsymbol{x_{k}} \mid \mathbb{Z}_{1:k-1})} {f( Z_{k} \mid \mathbb{Z}_{1:k-1})} $ 
\end{itemize}

% With the star-convex shape model, a GP can be adopted to model the joint distribution of the target spatial position and kinematic states \cite{granstrom2016extended}. 
In \cite{wahlstrom2015extended}, a method was presented in which the unknown radial function parametrizing the contour was modeled using a Gaussian Process (GP) while simultaneously estimating the kinematic state of the target. 
% In \cite{wahlstrom2015extended}, GPs are used for two different measurement models to describe the target contour and surface, respectively. 
% The authors used the GPs to model the star-convex shape radial function and measurements with respect to the target position and contour parameters. 
An Extended Kalman Filter(EKF) was employed as the estimator for the prediction and update iterations.

\section{Methodology and Experiment}

\subsection{Heuristic object detection}
To track a mobile robot on a factory floor with edge devices, we proposed an efficient heuristic object detection approach utilizing the prior information about the robot and environment. The details of the object detection are described in Algorithm \ref{object_detection}.

To remove the floor points, the ground plane in the sensor coordinate is defined as $ax+by+cz+d=0$, where the coefficients $\boldsymbol{m}$ are initially set for LiDAR and the camera in our case as $[0,0,1,1]^T$ and $[0,1,0,-0.5]^T$, respectively. In the initialization phase, points near the initial plane are selected and fed into RANSAC \cite{derpanis2010ransac} to obtain the optimal plane coefficients $\boldsymbol{m^*}$, which will be used for each subsequent point cloud frame. 
The used parameters are $\gamma_{0} = 1\ m$, $\gamma_{1} = 0.02\ m$, $\alpha = 0.1$, $MAX\_ITER=100$, $n = \ 10$.

In the detection phase, for each point cloud frame, ground points are removed based on their distance to the plane defined by the coefficients $\boldsymbol{m^*}$. Next, points outside the operation area are removed according to the predefined range information. Bounding boxes are then computed using Principal component analysis (PCA) \cite{dimitrov2009bounds} for each target candidate, which is a points cluster generated by DBSCAN. Subsequently, geometric features are extracted from the bounding box of each candidate. The geometric features used in this paper are the lengths of three orthogonal edges, their variance, and the areas of three orthogonal faces. The same types of geometric features for the target robot are computed from prior information such as the URDF \cite{sucan2019urdf} file. Finally, using these features, the candidate with the lowest detection cost, provided it is below the threshold, is considered the target robot, and the corresponding points are fed into the EOT framework. The used parameters are $\gamma_{2} = \ 0.02$, $\gamma_{3} = \ 1$, $\boldsymbol{f_p} = [0.39, 0.33, 0.21, 0.005, 0.13, 0.08, 0.07]^T$, $\boldsymbol{w_f} = [0.5, 0.5, 0.5, 100, 2, 1, 1]^T$, $\boldsymbol{m_c} = [0.03, 30]^T$. $\boldsymbol{m_{area}}$, for the camera, it is defined by the bounds $ -4 \leq x \leq4, -2 \leq y \leq 1, 0 \leq z \leq 2.5$; for the LiDAR, the bound is defined as $-4 \leq x \leq4, 0 \leq y \leq 2.7, -4 \leq z \leq 2$.

\begin{algorithm}[H]
% \begin{breakablealgorithm}
    \caption{Heuristic Object Detection}

    \label{object_detection}
        
    % \SetKwFunction{isOddNumber}{isOddNumber}

    // Initialization\;
    % \BlankLine
    % \SetKwInOut{KwIn}{Input}
    % \SetKwInOut{KwOut}{Output}
    \SetKwInOut{Input}{input}\SetKwInOut{Output}{output}
    % \KwIn{One frame of 3D point cloud $\boldsymbol{p} \in \mathbb{R}^{n\times3}$}
    % \KwOut{Processed list.}
    \Input{One point cloud $\boldsymbol{P_{0}} \in \mathbb{R}^{n\times3}$, initial ground coefficients $\boldsymbol{m} = [a, b, c, d]^T$, ground threshold $\gamma_{0}$, plane estimation threshold $\gamma_{1}$, down sample rate $\alpha$, MAX\_ITER, initial inlier number $n$}
    \Output{Optimal ground coefficients}

    \For{each point $\boldsymbol{p} \in \mathbb{R}^{3}$ in $\boldsymbol{P_{0}}$ }{
        calculate $dist = \lvert [a, b, c] \cdot \boldsymbol{p} \rvert$ \; 
        \eIf {$dist \leq \gamma_{0}$}{
            keep the $\boldsymbol{p}$
        }{
        drop the $\boldsymbol{p}$
        }
    }
    get the reduced point cloud $\boldsymbol{P}_{1} \in \mathbb{R}^{k\times 3}$ and down sample $\boldsymbol{P}_{2} = voxel\_down\_sample(\boldsymbol{P}_{2}, \alpha)$ \;
    estimate optimal coefficients with RANSAC: $\boldsymbol{m^{*}} = RANSAC(\boldsymbol{P}_{2}, n, MAX\_ITER, \gamma_{1})$
    
    \KwRet{$\boldsymbol{m^{*}}$ }
    \BlankLine
    
    // Density-based robot detection \;
    \Input{Point cloud at time step $k$ $\boldsymbol{P}_{0, k}$, ground coefficients $\boldsymbol{m^{*}}$, ground threshold $\gamma_{2}$, detection threshold $\gamma_{3}$,prior feature parameters $\boldsymbol{f_p}$, feature weights $\boldsymbol{w_f}$, operation area coefficients $\boldsymbol{m_{area}}$, clustering parameters $\boldsymbol{m_c}$}
    \Output{Robot points at time step $k$ $\boldsymbol{P}_{r, k}$}
    remove points out of operation area $\boldsymbol{P_{1, k}} = remove\_out\_of\_range\_points(\boldsymbol{P}_{0, k}, \boldsymbol{m_{area}})$ \;
    remove ground points $\boldsymbol{P}_{2, k} = remove\_ground\_points(\boldsymbol{P}_{1, k}, \gamma_{2})$ \;
    get clusters $\boldsymbol{C}_{k} = DBSCAN(\boldsymbol{P}_{2, k}, \boldsymbol{m_c})$ \;
    $\boldsymbol{P}_{r, k} \leftarrow \emptyset$, $l \leftarrow \gamma_3$ \;
    \For{each cluster $\boldsymbol{c} \in \mathbb{R}^{n \times 3}$ in\ $\boldsymbol{C}_{k}$}{
        compute the bounding box based on the PCA $bbox = get\_bounding\_box(\boldsymbol{c})$ \;
        extract geometric feature of the bounding box $\boldsymbol{f_{box}}$ \;
        compute the detection cost $J = \boldsymbol{w_f}^T \cdot \lvert \boldsymbol{f_{box}} - \boldsymbol{f_p} \rvert$ \;
        \If{$J \leq l$}{
        $\boldsymbol{P}_{r, k} \leftarrow \boldsymbol{c} $, $ l \leftarrow J$
        }
    }
    
    \KwRet{$\boldsymbol{P}_{r, k}$ }
\end{algorithm}

\subsection{Robot tracking}
The outputs from the object detection algorithm are used as measurements with which to perform extended object tracking. This tracking algorithm provides an estimate of the robot's spatial extent and velocity, along with the associated uncertainty for these quantities. We use the previously mentioned GP model with a constant velocity motion model and employ an iterated EKF to handle the non-linearity, as done in \cite{baerveldt2023extended}. Since this model assumes contour-generated measurements, we first remove all measurements that do not originate from the contour. To initialize the estimate, we use the dimensions of the robot to define the prior extent.
The GP model requires parameters to specify the GP covariance function as well as measurement and process noise. In this work, we use $\sigma_c$ = 0.05 $\sqrt{m^2 s^{-3}}$ as the noise parameter for the constant velocity model and $\sigma_r$ = 0.1 m for the measurement noise. We use 10 test angles to parametrize the extent and the GP covariance function hyperparameters are $\sigma_f$ = 0.01 m, $\sigma_r$ = 0.005 m, $\sigma_n$ = 0.001 m, $l=\pi/6$ rad and the forgetting factor $\eta_f=0.001$. For information on how these parameters are defined, see \cite{baerveldt2023extended}.

\subsection{Setup and results}
\begin{figure}[h]
    \centering
    \begin{minipage}{0.3\textwidth}
        \centering
        \includegraphics[width=\textwidth]{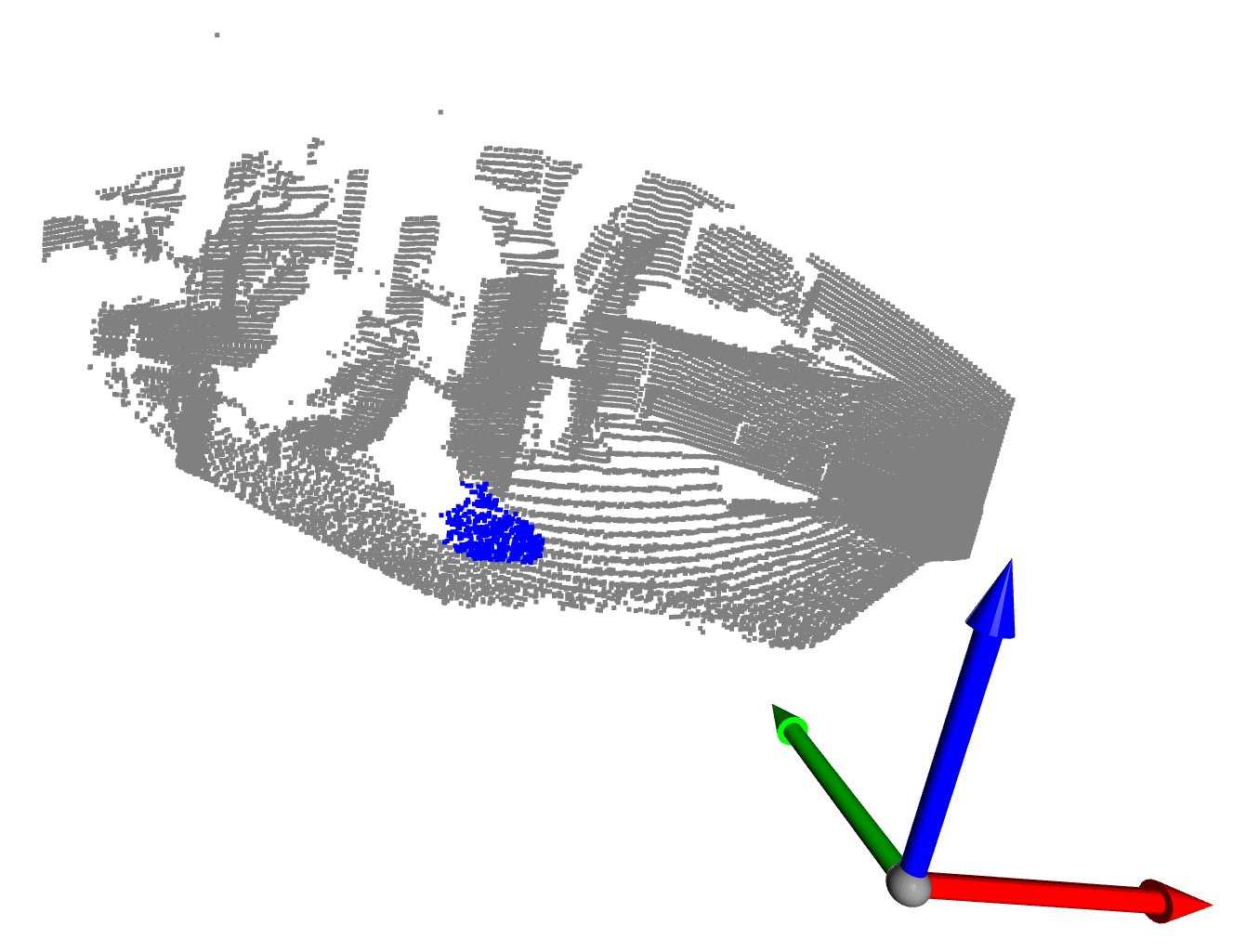}
        \caption{Point cloud from Cube}
        \label{fig:lidarpcd}
    \end{minipage}\hfill
    \begin{minipage}{0.3\textwidth}
        \centering
        \includegraphics[width=\textwidth]{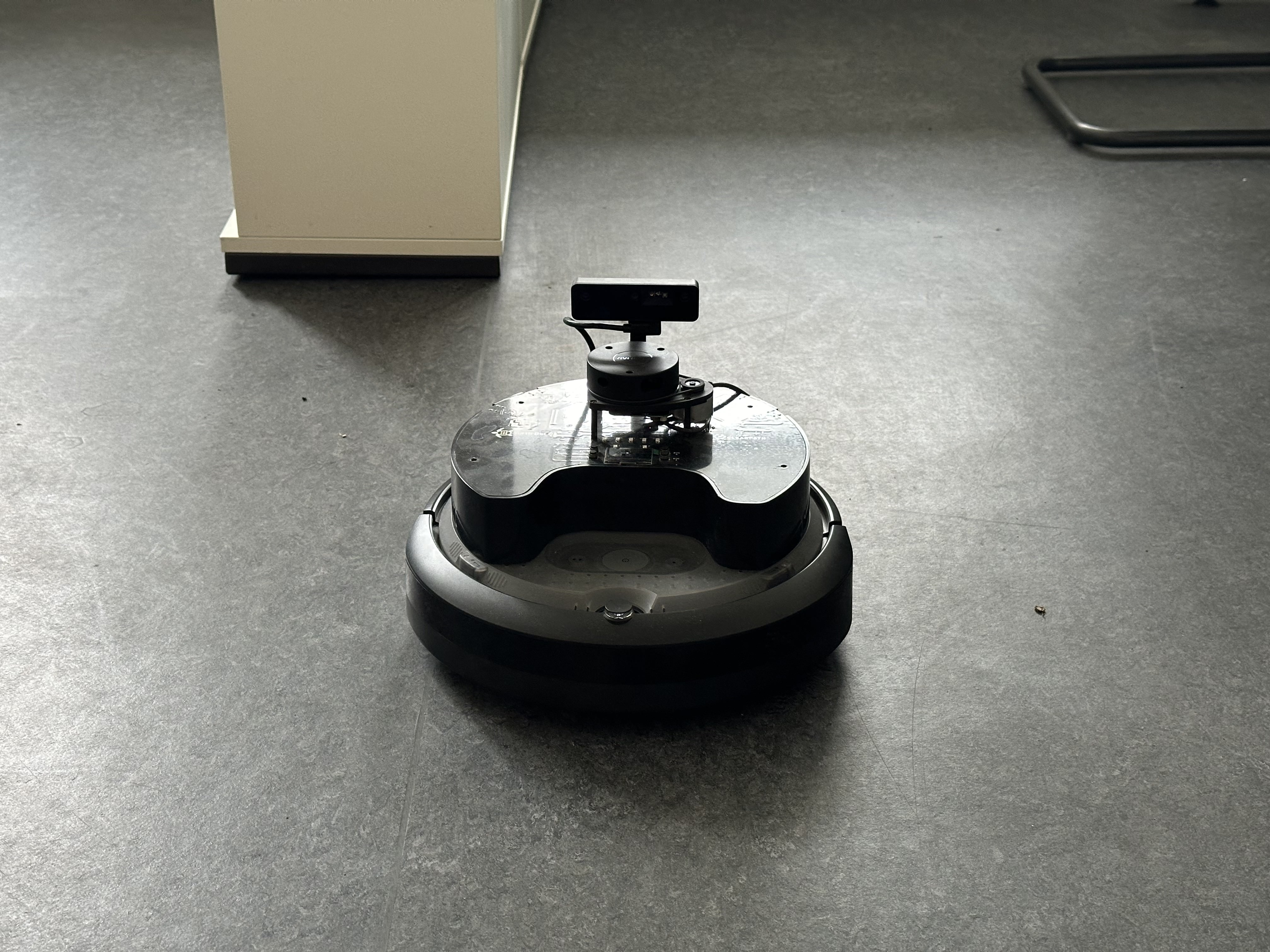}
        \caption{TurtleBot 4}
        \label{fig:turtlebot}
    \end{minipage}\hfill
    \begin{minipage}{0.3\textwidth}
        \centering
        \includegraphics[width=\textwidth]{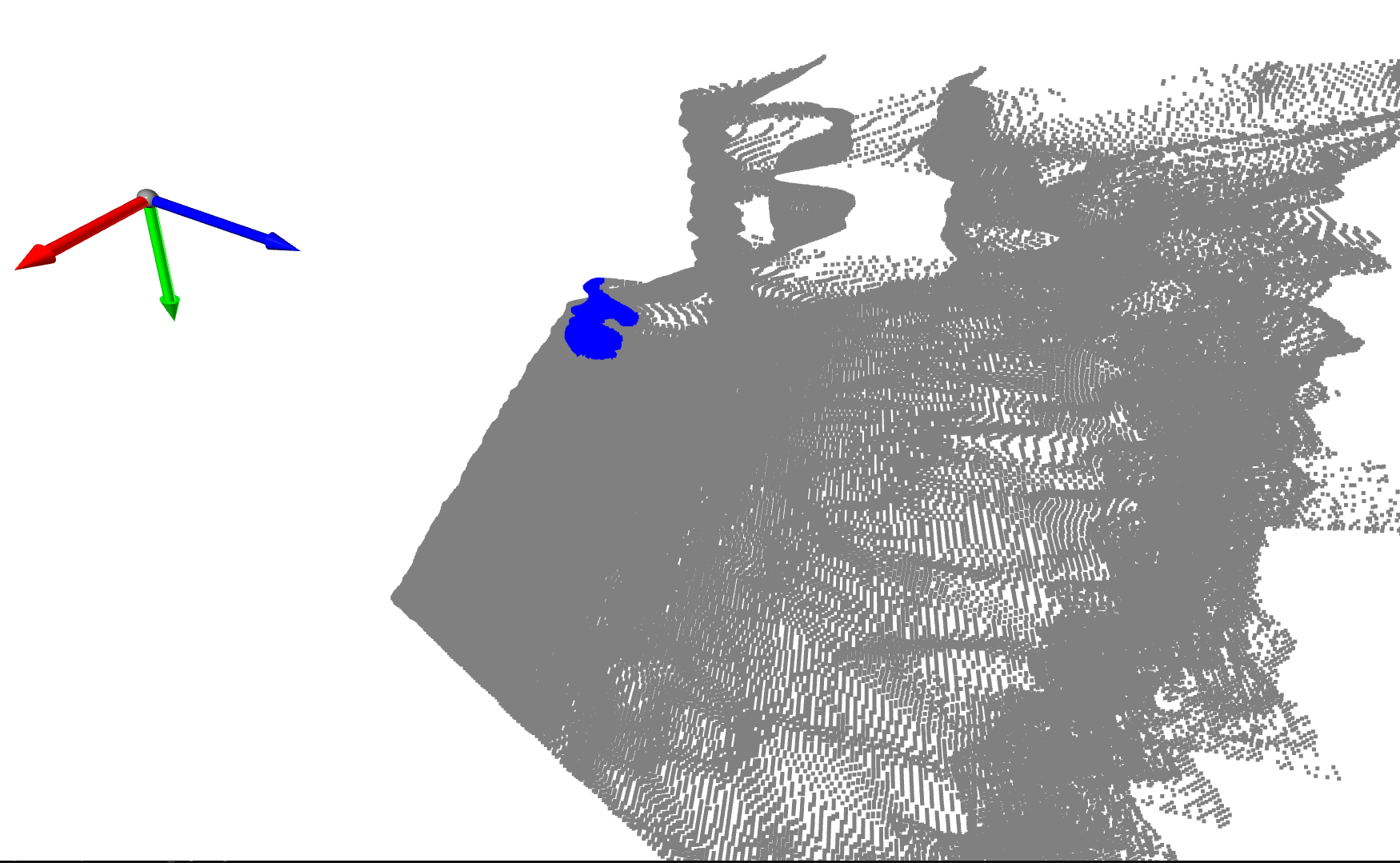}
        \caption{Point cloud from D435i}
        \label{fig:camerapcd}
    \end{minipage}
    \label{fig:overall}
\end{figure}

We evaluated our method in an indoor environment using the TurtleBot 4 \cite{turtlebot4} as shown in Figure \ref{fig:turtlebot}, with the Blickfeld Cube 1 and Intel Realsense D435i as the selected sensors. The Cube 1 can operate at a distance range of 1.5 to 75 meters with an error within 2 cm \cite{blickfeld}, while the D435i can operate at a distance range of 0.3 to 3 meters with a depth error of less than $2\%$ \cite{d435i}.
We collected two motion trajectories of the robot with both sensors within a range of 3 meters: one trajectory involved straight movement, while the other included turns. The LiDAR point cloud frames were collected at 4.4 Hz, and the camera point cloud frames at 30 Hz.

\begin{figure}[h]
    \centering
    \begin{subfigure}[b]{0.3\textwidth}
        \centering
        \includegraphics[width=\textwidth]{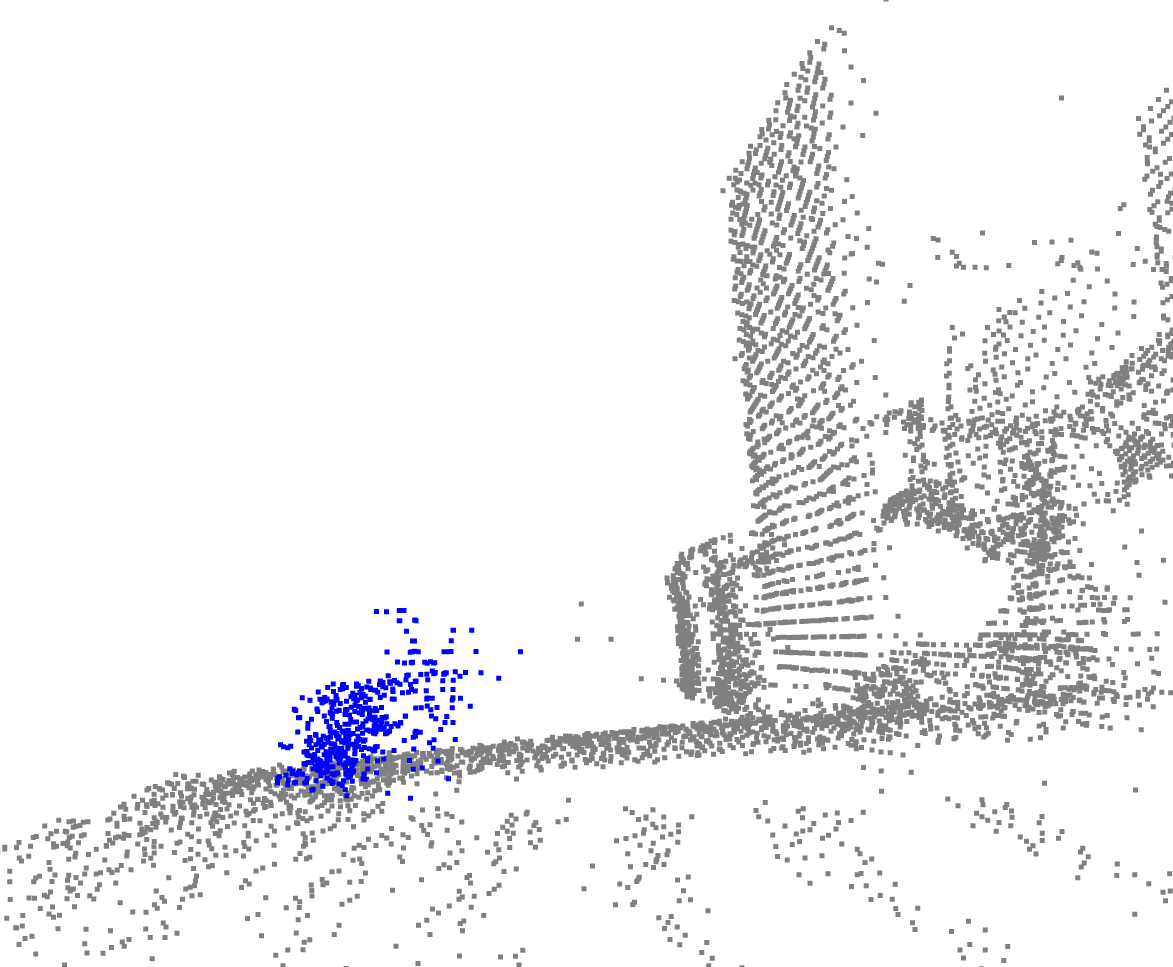}
        \caption{Cube LiDAR}
        \label{fig:cube_res}
    \end{subfigure}
    \hfill
    \begin{subfigure}[b]{0.3\textwidth}
        \centering
        \includegraphics[width=\textwidth]{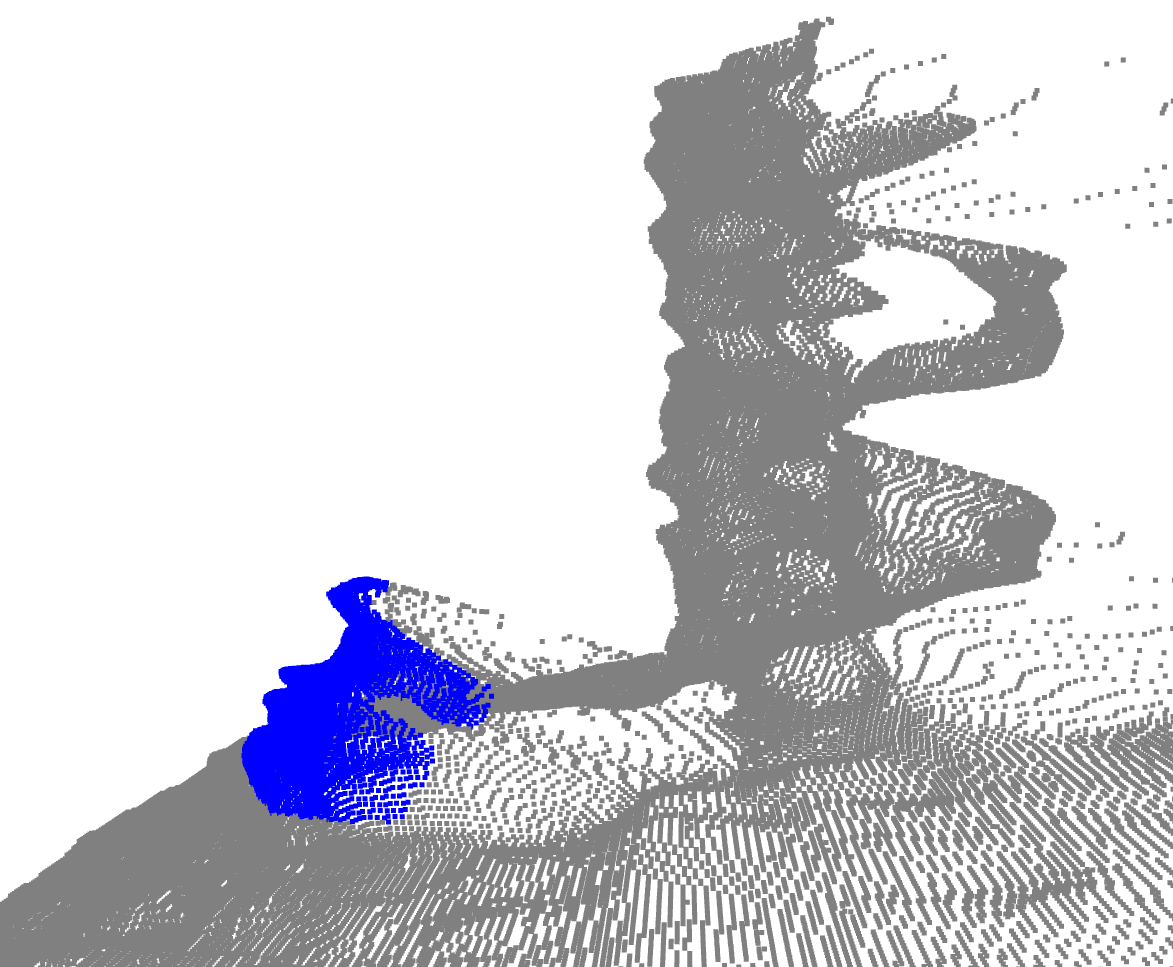}
        \caption{D435i camera}
        \label{fig:d435_res}
    \end{subfigure}
    \caption{Detected robot points (blue) in point clouds of two sensors.}
    \label{fig:two_detection}
\end{figure}

Figures \ref{fig:lidarpcd}, \ref{fig:camerapcd}, \ref{fig:two_detection} depict the results of the proposed object detection algorithm on LiDAR and camera point clouds. Gray points in the figures represent background points, while blue points denote detected robot points. The coordinate axes in the figures are colored red, green, and blue to indicate the XYZ axes in the Cartesian coordinate systems of the respective sensors. Upon comparing Figure \ref{fig:d435_res} with Figure \ref{fig:cube_res}, and Figure \ref{fig:camerapcd} with Figure \ref{fig:lidarpcd}, it is evident that the point cloud from the D435i sensor exhibits higher levels of noise, particularly pronounced at object edges and occluded regions from the camera's viewpoint. Furthermore, the noise distribution intensifies with increasing distance from the camera. Our proposed object detection algorithm has achieved effective detection performance in both types of point clouds: achieving a detection rate of $98\%$ in LiDAR point clouds and $91\%$ in camera point clouds.

\begin{figure}[h]
    \centering
    \begin{subfigure}[b]{0.48\textwidth}
        \centering
        \includegraphics[width=1.1\textwidth]{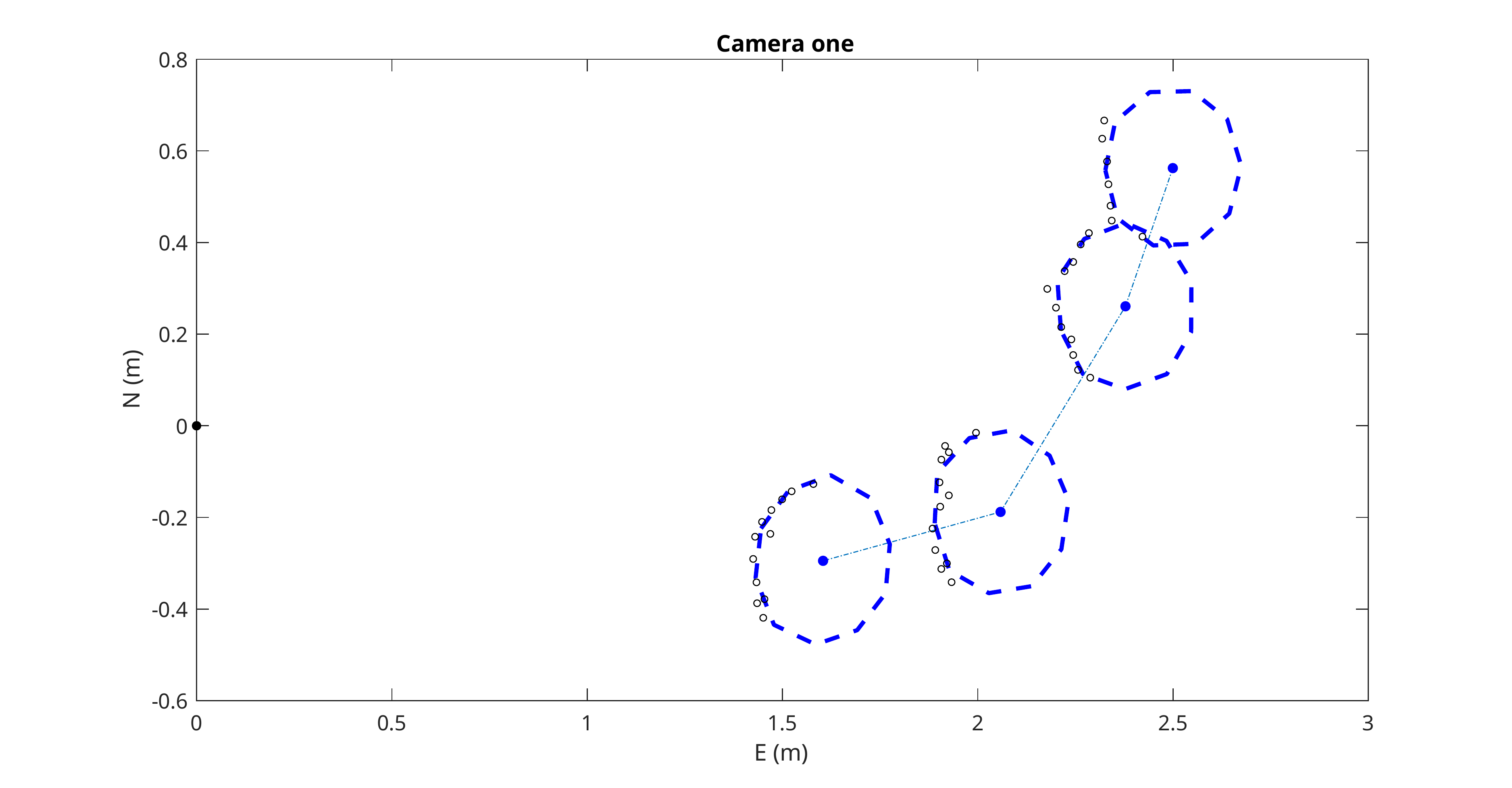}
        \caption{Tracking turning motion with camera}
        \label{fig:traj_1_camera}
    \end{subfigure}
    \hfill
    \begin{subfigure}[b]{0.48\textwidth}
        \centering
        \includegraphics[width=1.1\textwidth]{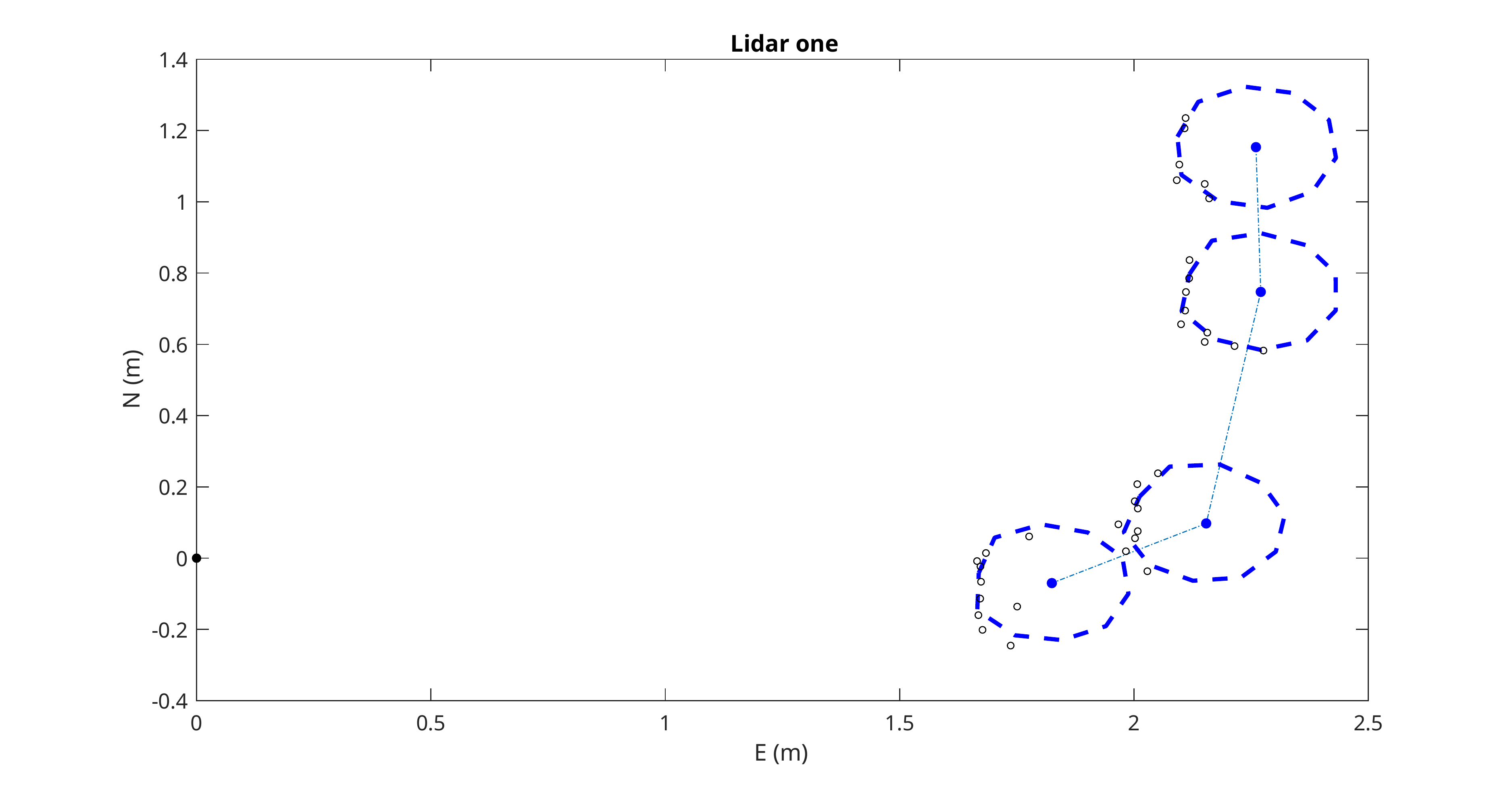}
        \caption{Tracking turning motion with LiDAR}
        \label{fig:traj_1_lidar}
    \end{subfigure}
    \vskip\baselineskip
    \begin{subfigure}[b]{0.48\textwidth}
        \centering
        \includegraphics[width=1.1\textwidth]{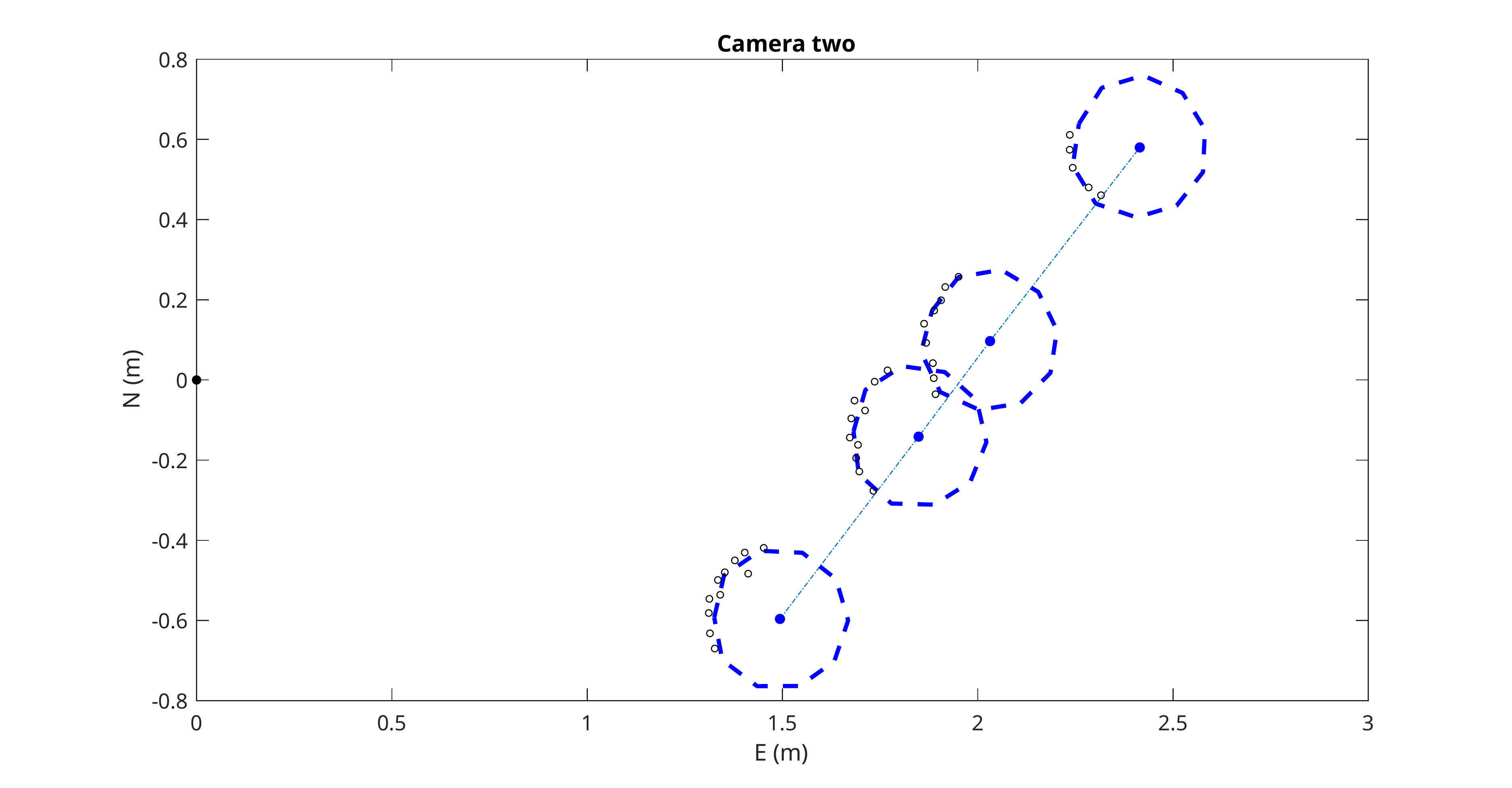}
        \caption{Tracking straight motion with camera}
        \label{fig:traj_2_camera}
    \end{subfigure}
    \hfill
    \begin{subfigure}[b]{0.48\textwidth}
        \centering
        \includegraphics[width=1.1\textwidth]{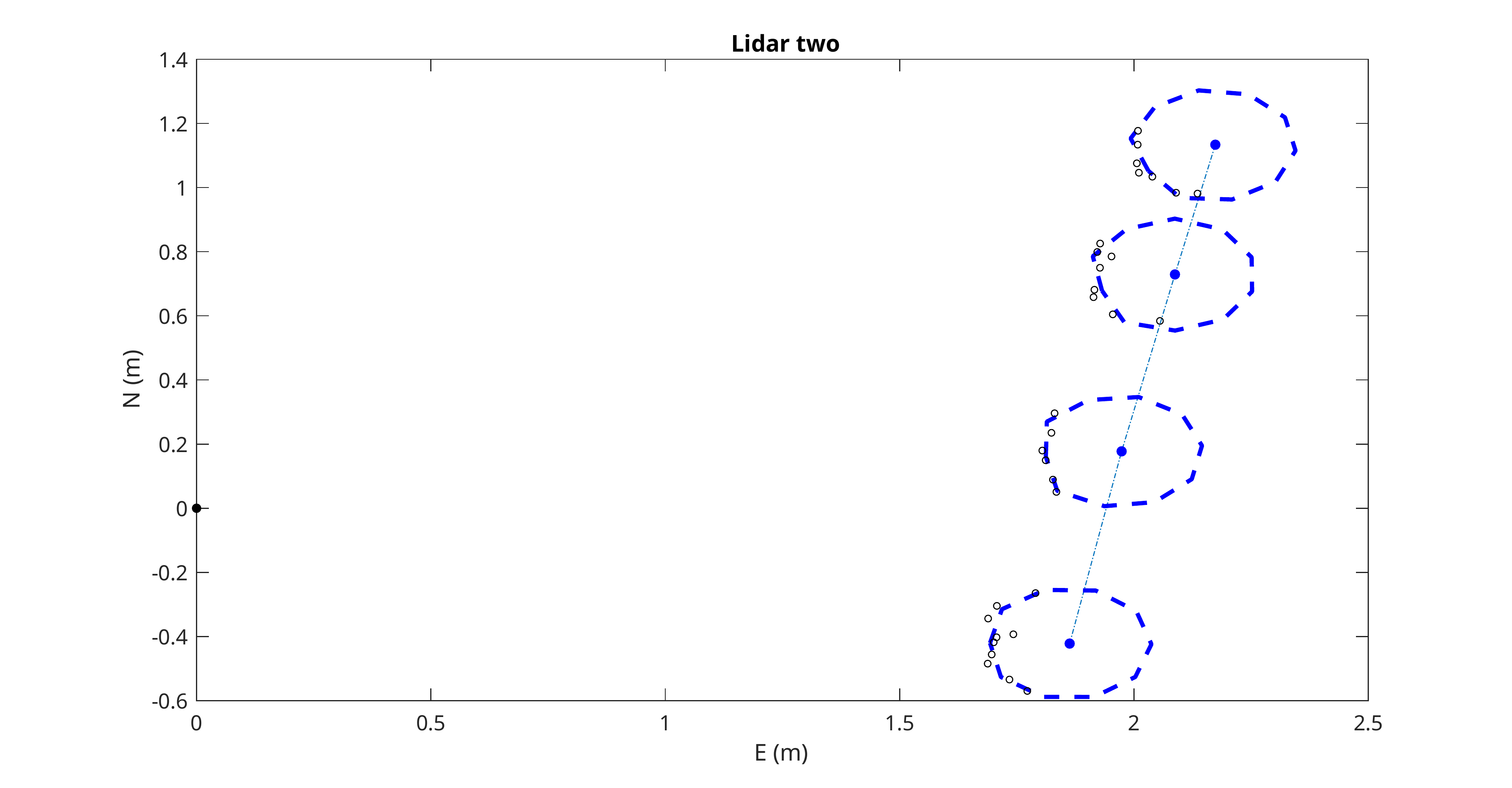}
        \caption{Tracking straight motion with LiDAR}
        \label{fig:traj_2_lidar}
    \end{subfigure}
    \caption{Tracking results of two motion trajectories with Cube LiDAR and D435i camera. The coordinate in each image is on a Cartesian plane of the corresponding sensor's local coordinates. The light blue curve in each image represents the motion trajectory of the estimated centroid. Four representative timesteps' results are visualized in each image: the blue dashed line represents the estimated shape, the blue dot represents the centroid, and the black circles represent the sensor measurements.}
    \label{fig:four_images}
\end{figure}

Figure \ref{fig:four_images} demonstrates that our method successfully achieved similar tracking performance for both types of trajectories across two different sensors. The robot's extent was fairly captured, effectively reflecting the characteristics of each movement, including straight segments and turns.
Notably, due to the differing poses of the two sensors in the world frame and the visualization results being directly projected onto the Cartesian plane of each sensor's coordinate system, there are visual shape differences caused by distortion.

\subsection{Discussions and Open Research Challenge}

In this preliminary work, the proposed tracking method is highly dependent on its heuristic object detector. The detector comprises multiple hyperparameters, necessitating extensive fine-tuning in complex and dynamic environments. Moreover, within this study, these hyperparameters are not adjustable during operation. Through our analysis of point clouds obtained from the camera, we observed that noise distribution is strongly correlated with depth. Consequently, when the target's trajectory exhibits significant variations in the depth dimension, the fixed hyperparameters make it challenging for the detector to filter out noise effectively. Introducing a parameter updating mechanism into the detector, such as adaptive DBSCAN\cite{el2022adaptive} or similar approaches like filter banks\cite{lin2021real} to establish a feasible parameter set, may provide a viable solution. Furthermore, in our employed EOT framework, the measurement noise model did not account for the correlation with depth. In future work, we intend to incorporate depth-related measurement noise models into the EOT framework.

\section{Conclusion}
In this work, we conducted a preliminary performance comparison of single object tracking on 3D point clouds from a LiDAR and a stereo camera using the proposed method, which consists of a heuristic object detector and an EOT framework. We evaluated our method in an indoor environment by tracking a mobile robot with a Blickfeld Cube 1 LiDAR and an Intel Realsense D435i stereo camera, which have a significant price difference. The results demonstrate that our method achieves similar performance on both sensors, indicating the feasibility of using EOT with stereo cameras. In the future, we plan to evaluate the tracking results using ground truth data from the robot’s onboard sensors and to explore adaptive methods for the detector and a depth-correlated noise model for the EOT framework.

\begin{acknowledgments}
  This work was funded by the Horizon Europe Research
and Innovation Actions under grant number 101092908
(SmartEdge).
  The co-author Martin Baerveldt is a researcher funded by the European
    Union’s Horizon 2020 research and innovation programme under the Marie
    Skłodowska-Curie grant agreement No 955.768 (MSCA-ETN AUTOBarge).
  
\end{acknowledgments}

%%
%% Define the bibliography file to be used
\bibliography{sample-ceur}

%%
% %% If your work has an appendix, this is the place to put it.
% \appendix


% The sources for the ceur-art style are available via
% \begin{itemize}
% \item \href{https://github.com/yamadharma/ceurart}{GitHub},
% % \item \href{https://www.overleaf.com/project/5e76702c4acae70001d3bc87}{Overleaf},
% \item
%   \href{https://www.overleaf.com/latex/templates/template-for-submissions-to-ceur-workshop-proceedings-ceur-ws-dot-org/pkfscdkgkhcq}{Overleaf
%     template}.
% \end{itemize}

\end{document}